\documentclass[10pt,twocolumn,letterpaper]{article}

\usepackage[pagenumber]{cvpr}           % To produce the CAMERA-READY version

\usepackage[accsupp]{axessibility}
\usepackage{times}
\usepackage{epsfig}
\usepackage{graphicx}
\usepackage{amsmath}
\usepackage{amssymb}
\usepackage{dsfont}
\usepackage{amsthm}
\usepackage{mathtools}
\usepackage[numbers,sort]{natbib}
\usepackage[pagebackref,breaklinks,colorlinks]{hyperref}

\usepackage[capitalize]{cleveref}
\crefname{section}{Sec.}{Secs.}
\Crefname{section}{Section}{Sections}
\Crefname{table}{Table}{Tables}
\crefname{table}{Tab.}{Tabs.}

\def\cvprPaperID{8328} % *** Enter the CVPR Paper ID here
\def\confName{CVPR}
\def\confYear{2022}

\newtheorem{axiom}{Axiom}

\newtheorem{property}{Property}
\theoremstyle{definition}
\newtheorem{condition}{Condition}[section]

\begin{document}

%%%%%%%%% TITLE
\title{AxIoU: An Axiomatically Justified Measure for Video Moment Retrieval}

\author{Riku Togashi\\
Cyberagent, Inc., Waseda University
\and
Mayu Otani\\
Cyberagent, Inc.
\and
Yuta Nakashima\\
Osaka University
\and
Esa Rahtu\\
Tampere University
\and
Janne Heikkil\"{a}\\
University of Oulu
\and
Tetsuya Sakai\\
Waseda Univeristy
}

\maketitle

% %%%%%%%%% ABSTRACT
\begin{abstract}
  Evaluation measures have a crucial impact on the direction of research.
  Therefore, it is of utmost importance to develop appropriate and reliable evaluation measures for new applications where conventional measures are not well suited.
  Video Moment Retrieval (VMR) is one such application, and the current practice is to use R@$K,\theta$ for evaluating VMR systems.
  However, this measure has two disadvantages.
  First, it is rank-insensitive: It ignores the rank positions of successfully localised moments in the top-$K$ ranked list by treating the list as a set.
  Second, it binarizes the Intersection over Union (IoU) of each retrieved video moment using the threshold $\theta$ and thereby ignoring fine-grained localisation quality of ranked moments.
  
  We propose an alternative measure for evaluating VMR, called Average Max IoU (AxIoU), which is free from the above two problems.
  We show that AxIoU satisfies two important axioms for VMR evaluation, namely, 
  \textbf{Invariance against Redundant Moments}
  and 
  \textbf{Monotonicity with respect to the Best Moment}, and also that R@$K,\theta$ satisfies the first axiom only.
  We also empirically examine how AxIoU agrees with R@$K,\theta$, as well as its stability with respect to change in the test data and human-annotated temporal boundaries.
\end{abstract}

\section{Introduction}
Video Moment Retrieval (VMR) has been explored to find relevant fragments of videos (\ie video moments) based on a user's textual query~\cite{gao2017tall,hendricks17iccv}.
Most existing VMR systems \cite{gao2017tall,wu2018multi,liu2018attentive,zhang2019man,yuan2019semantic} cast the problem of finding video moments into a ranking problem.
For evaluating ranked lists of video moments, R@$K,\theta$ is widely adopted in the literature \cite{gao2017tall}.
R@$K,\theta$ for a query $q$ is defined as 1 if at least one relevant video moment in the top $K$ of the ranked list has an Intersection over Union (IoU) larger than $\theta$ with the ground truth for $q$.

\begin{figure}[t!]
    \centering
    \includegraphics[clip,width=\linewidth]{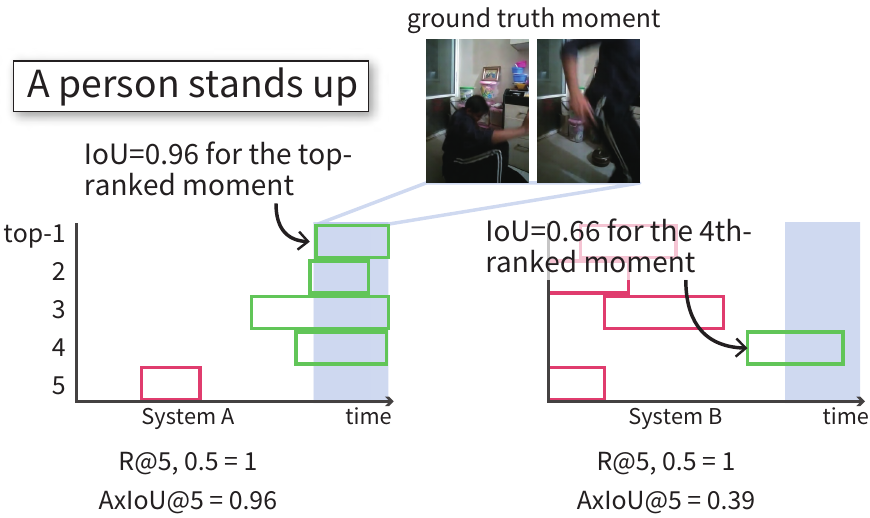}
    \caption{
    The system on the left shows a moment with a large overlap with the ground truth (blue band) in the top of the ranked list, and the system on the right illustrates a moment with a much smaller overlap with the ground truth at rank 4.
    According to R@$5,0.5$, the two systems are equally effective.
    We propose AxIoU whose measurements reflect the localisation quality (\ie IoU) and the rank of successfully retrieved video moments.
    The photo in the figure is taken from Charades-STA~\cite{gao2017tall}.}
    \label{fig:overview}
\end{figure}

R@$K,\theta$ has two disadvantages as illustrated in Figure~\ref{fig:overview}.
First, it is \textit{rank-insensitive}, as the video moments in the top-$K$ ranked list are treated as a set, and their ranks are not considered.
Second, it is \textit{localisation-insensitive}, \ie, the exact position (start and end points) of the video moment does not affect the measurement as it binarizes the IoU of each video moment using threshold $\theta$.
Thus, R@$K,\theta$ only provides a binary measurement for a ranked list in an all-or-nothing manner, ignoring the ranking and localisation quality of top-$K$ predicted video moments.
As we shall demonstrate in this paper, these properties of R@$K,\theta$ are problematic for reliable evaluation.
R@$K,\theta$ cannot distinguish ranked lists with different quality due to the binary property, while leading to instability under a small number of evaluation samples and label ambiguity~\cite{sigurdsson2017actions,hendricks17iccv,alwassel2018diagnosing,otani2020challengesmr}.
Moreover, R@$K,\theta$ evaluates rather different aspects of system quality depending on a parameter setting and can conflict with each other;
for example, a ranked list whose second moment achieves $\mathrm{IoU}=0.71$ is measured to be $1.0$ in terms of R@$2,0.7$ but to be $0.0$ in terms of R@$1,0.7$ when the first moment has $\mathrm{IoU}=0.69$.
These undesirable properties of R@$K,\theta$ should be carefully considered for future studies because conclusions drawn from potentially unstable measures may not generalise well.
In practice, the instability of R@$K,\theta$ implies that 
we may underestimate a VMR method by adopting a non-best model based on validation R@$K,\theta$.

In this paper, we propose an alternative measure for evaluating VMR systems, called \emph{Average Max IoU} (AxIoU), which does not suffer from the problems above with R@$K,\theta$.
To evaluate evaluation measures, 
we take an axiomatic approach \cite{fang2011diagnostic,sebastiani2015axiomatically,amigo2018axiomatic,sebastiani2020evaluation} and introduce two important axioms that an effectiveness measure for VMR must satisfy, 
namely, 
\emph{invariance against redundant moments}
and \emph{monotonicity with respect to the best moment}.
We show that R@$K, \theta$ only satisfies the first axiom. 
We also empirically investigate the properties of AxIoU
in practical terms, namely, agreement with conventional R@$K,\theta$ and the stability to the size of a dataset and label ambiguity.

\section{Related Work\label{related_work}}
Most of the prior studies of VMR adopted the R@$K,\theta$ measure~\cite{gao2017tall,krishna2017dense,wu2018multi,liu2018attentive,zhang2019man,yuan2019semantic,escorcia2019temporal,Zeng_2020_CVPR,gao2021fast,wang2021structured}.
Gao \etal \cite{gao2017tall} proposed the use of this measure for VMR by referring to the work of Hu \etal~\cite{hu2016natural}, which is an early study on an object retrieval task with textual queries.
The values of $K$ and $\theta$ are chosen for each dataset.
For example, the combinations of $K=1,5,10$ and $\theta=0.3,0.5,0.7$ are widely adopted in Charades-STA~\cite{gao2017tall}, ActivityNet~\cite{caba2015activitynet,krishna2017dense}, and DiDeMo~\cite{hendricks17iccv}.
In the TACoS dataset~\cite{regneri2013grounding}, relatively relaxed values of $\theta$ (\ie $\theta=0.1,0.3,0.5$) are used.
For evaluating methods that output only one moment per query \cite{he2019read,ghosh-etal-2019-excl,Yuan_Mei_Zhu_2019},
R@$K,\theta$ with $K=1$ is often adopted.
Lei \etal recently proposed a new retrieval task called video corpus moment retrieval~\cite{lei2020tvr}, in which
a system requires to retrieve relevant moments from multiple videos.
Owing to a large number of candidate moments, they utilise large values of $K$ such as $K=100$.
However,
the common practice of reporting multiple settings of R@$K,\theta$ is controversial.
As we shall demonstrate in this paper, different parameter settings often lead to different system rankings, from which it may be difficult to draw useful conclusions from the evaluation.

Prior studies suggested that the inter-rater agreement of human-annotated temporal boundaries is often not strong~\cite{sigurdsson2017actions,hendricks17iccv,alwassel2018diagnosing,otani2020challengesmr}.
Hendricks \etal found that there are multiple video moments, which can be described by a textual query~\cite{hendricks17iccv}; to alleviate this label ambiguity, they developed a user interface.
Sigurdsson \etal and Alwassel \etal also reported that human-annotated temporal regions do not agree well with each other~\cite{sigurdsson2017actions,alwassel2018diagnosing}.
Otani \etal observed high label ambiguity in Charades-STA and ActivityNet~\cite{otani2020challengesmr}.
Nevertheless, the binarization of IoU values in R@$K,\theta$ introduces potential instability to the change of labels.
In particular, a large value of $\theta$ requires exactly located temporal regions and thereby being noisy, inheriting label ambiguity.

In the context of object detection, in which evaluation measures often rely on a threshold parameter for spatial IoU,
prior studies have discussed the disadvantages of a fixed threshold~\cite{LRP-ECCV18,hall2020probability,electronics10030279}.
On the MSCOCO~\cite{lin2014microsoft} dataset,
an average of measures over IoU threshold values is adopted for evaluating fine-grained localisation quality; the measure is called COCO mean average precision (mAP).
Oksuz \etal proposed an object detection measure to directly quantify the bounding box tightness by introducing IoU values without thresholding in their measure~\cite{LRP-ECCV18}.
Hall \etal have explored a way to improve the spatial quality evaluation of detected regions beyond the conventional box-based IoU while reducing the parameters in evaluation measures~\cite{hall2020probability}.
In temporal localisation tasks for videos (\eg action detection), Alwassel \etal used a COCO mAP-like measure for evaluation~\cite{alwassel2018diagnosing}.

In contrast to rank-insensitive set retrieval measures (\eg precision and recall),
ranked retrieval measures have been explored for evaluating the quality of a list of ranked items, such as normalised discounted cumulative gain (nDCG)~\cite{jarvelin2002cumulated}.
Such measures often have weights for the rank positions in a retrieval result;
for example, the discount function in nDCG can be regarded as the importance of each position.
Based on the interpretation of the position weights from the viewpoint of \emph{user models},
prior studies have developed various evaluation measures~\cite{moffat2008rank,robertson2008new,sakai2008modelling,chapelle2009expected}.

The evaluation of evaluation measures is often challenging as it requires the true evaluation results a priori.
One approach to verify the experiments based on an evaluation measure is to collect human manual assessments for \emph{search engine result pages (SERPs)}~\cite{sanderson2010user}.
For new applications such as VMR, it is often costly to establish a reliable environment to collect the gold data that aligns well with the ``true'' quality; we may need to study such as human effects on the reliability of the gold data~\cite{kazai2013analysis}.
An axiomatic approach is another direction for the verification of evaluation measures~\cite{fang2011diagnostic,sebastiani2015axiomatically,amigo2018axiomatic,sebastiani2020evaluation}.
By formally defining requirements that a measure should satisfy, we can analytically confirm the validity of measures.
Such requirements inevitably depend on a number of assumptions.
However, this is also true for the assessment-based approach because guidelines for assessors implicitly involve assumptions on users' behaviours~\cite{sanderson2010user}.

In this paper,
we propose an alternative VMR measure, AxIoU, which is an instantiation of normalised cumulative utility (NCU)~\cite{sakai2008modelling,sakai2013metrics}, which is a wide class of information retrieval measures including AP.
Our proposed measure considers the rank positions and IoU values of video moments.
To confirm the properties of measures, we take an axiomatic approach.
The derivation of AxIoU is related to COCO mAP, whereas AxIoU analytically reduces the binarization process for IoU values.
Through empirical experiments, we confirm the numerical properties of AxIoU while showing the undesirable behaviours of R@$K,\theta$. 

\section{Preliminaries}
\subsection{Notations}
Our goal is to develop a measure $\mu(q, \sigma)$
that estimates the retrieval effectiveness of a system $\sigma$
based on a test query $q$.
We also denote by $\mu(\mathcal{Q},\sigma)=(1/|\mathcal{Q}|)\sum_{q \in \mathcal{Q}}\mu(q, \sigma)$ the mean of the measurements based on a test query set $\mathcal{Q}$.
For a query $q \in \mathcal{Q}$, 
the system $\sigma$ sorts the set $\mathcal{M}_q$ of candidate moments and creates the ranked list $\sigma_q$.
We also denote by $\sigma_q(k) \in \mathcal{M}_q$ the moment ranked at position $k$ in $\sigma_q$.
Let $r_{q}(m) \in [0,1]$ be the relevance score of a moment $m \in \mathcal{M}_q$,
computed as the temporal IoU (Intersection over Union) between 
$m$ and the ground truth region for $q$. Where there is no ambiguity, we will also denote it by $r(m)$.

\subsection{R@$K,\theta$\label{section:conventional_measure}}
First, we formally define the conventional measure, R@$K,\theta$~\cite{gao2017tall}, and clarify what it quantifies as well as its limitations.
Here, we denote by $\mathds{1} \colon \mathbb{B} \mapsto \{0,1\}$ the indicator function for the boolean variable $X$ that takes 1 if $X$ is true and 0 if $X$ is false. 
We express R@$K,\theta$ and Mean R@$K, \theta$ as follows.
\begin{align}
\notag
  \text{R@}K,\theta(q, \sigma)
  &\coloneqq \mathds{1}\left\{\sum_{k=1}^{K}\mathds{1}\left\{r(\sigma_{q}(k)) > \theta\right\} > 0\right\} \\\label{eq:definition_recall}
  &= \mathds{1}\left\{\max_{1\leq k \leq K}r(\sigma_{q}(k)) > \theta \right\}.
\end{align}
The value of R@$K,\theta$ depends entirely on whether
the most relevant moment in the top-$K$ retrieved results exceeds the $\theta$ threshold. 
It is clear from this that R@$K, \theta$ does not reward redundancy:
the retrieved moments other than the most relevant one in the SERP do not count, even if they also exceed $\theta$.
We shall refer to such relevant moments as \emph{redundant} moments.

The above property of R@$K,\theta$ is a desirable feature, since real VMR system users probably do not care about redundant moments in their SERPs.
However, it is clear from Eq.~\ref{eq:definition_recall} that R@$K, \theta$ has two potential shortcomings.
First, R@$K,\theta$ is unchanged by the rank positions of the relevant moments: it is a set retrieval measure rather than a ranked retrieval measure.
For $K>1$, it cannot distinguish between a system that retrieves a perfectly relevant moment at rank~1,
and a system that retrieves the same moment at rank~$K$.
Second, it binarizes the IoU of each moment using the $\theta$ threshold, and thereby ignores the degree of relevance of each retrieved moment.
Choosing an appropriate value of $\theta$ is practically problematic, especially given that $K$ also needs to be chosen at the same time.

\section{Proposed Measure}
\subsection{Average Max IoU Measure\label{section:axiou_definition}}
To design a measure for VMR,
we adopt the framework of a wide class of retrieval effectiveness measures, \emph{normalised cumulative utility (NCU)} \cite{sakai2013metrics,sakai2008modelling}.
NCU assumes that there is a population of users who scan a ranked list, starting from the top, and abandons the ranked list on a certain rank position $k$.
Here, NCU for a query $q$ and a system $\sigma$ can be expressed as follows:
\begin{eqnarray}
  \text{NCU}(q, \sigma) = \sum_{k=1}^{|\mathcal{M}_{q}|} P_A(k)U(\sigma_q, k),
\end{eqnarray}
where $P_{A}(k)$ is the abandonment probability at rank position $k$ (\ie the population of users who stop at $k$), and
$U(\sigma_q, k)$ is the utility of the ranked list $\sigma_q$ at $k$.
As we do not want to reward redundancy in VMR, we follow the approach of R@$K,\theta$ (Eq.~\ref{eq:definition_recall}) to instantiate our utility function:
\begin{eqnarray}
  U(\sigma_q, k) = \max_{1 \leq j \leq k}r(\sigma_{q}(j)).
\end{eqnarray}
Based on this, we can obtain \emph{normalised cumulative max IoU (NCxIoU)} measure as follows:
\begin{eqnarray}
  \text{NCxIoU}(q, \sigma) \coloneqq \sum_{k=1}^{|\mathcal{M}_{q}|} P_A(k)\max_{1 \leq j \leq k}r(\sigma_{q}(j)).
\end{eqnarray}
With VMR, we do not have any prior knowledge on $P_A(k)$.
Therefore, given a SERP containing $K$ moments, we assume that the users are uniformly distributed over the $K$ moments:
that is, that $1/K$ of the user population abandons the list at rank $k$ ($1 \leq k \leq K$).
Note that the Average Precision (AP),
an NCU measure widely used in information retrieval evaluation with manual relevance assessments,
assumes that the users are uniformly distributed over \emph{all relevant} documents~\cite{robertson2008new}.
In the case of VMR, we consider only the top-$K$ items (following R@$K,\theta$),
and assume that each retrieved moment is at least somewhat relevant, where the degree of relevance is represented by the IoU of each moment.

Our proposed measure for VMR is also an instantiation of NCU, which we call 
average max IoU (AxIoU):
\begin{align}
  &\text{AxIoU@}K(q, \sigma) \coloneqq \frac{1}{K}\sum_{k=1}^{K}\max_{1 \leq j \leq k}r(\sigma_{q}(j)).
\end{align}
As the uniform assumption on $P_A(k)$ may not hold when with a large $K$,
we can use a more realistic distribution for $P_A(k)$ such as
the expected reciprocal rank (another NCU measure) \cite{chapelle2009expected},
although we leave this as future work.

\subsection{Interpretation of the AxIoU Measure\label{section:axiou_interpretation}}
In this section, we describe the relationship between our proposed measure and R@$K,\theta$.
We first consider the marginalisation of R@$K,\theta$ in terms of $K$ and $\theta$.
In practical terms, because we do not have any knowledge regarding the distribution of $\theta$ for each dataset, each query, or each set of systems to be evaluated,
we assume that $\theta \sim \text{Uni}(0, 1)$ and then obtain the following equation:
\begin{align}
  \notag
  & \mathbb{E}_{k}\mathbb{E}_{\theta}\left[\frac{1}{|\mathcal{Q}|}\sum_{q \in \mathcal{Q}}\text{R}@k,\theta(q,\sigma)\right] \\ \notag
  &=\mathbb{E}_{k}\mathbb{E}_{\theta}\left[\frac{1}{|\mathcal{Q}|}\sum_{q \in \mathcal{Q}}\mathds{1}\left\{\max_{1\leq j \leq k}r(\sigma_{q}(j)) > \theta \right\} \right]\\ \notag
  &=\frac{1}{|\mathcal{Q}|}\sum_{q \in \mathcal{Q}}\mathbb{E}_{k}\mathbb{E}_{\theta}\left[\mathds{1}\left\{\max_{1\leq j \leq k}r(\sigma_{q}(j)) > \theta \right\} \right]\\\label{eq:remove_theta}  
  &=\frac{1}{|\mathcal{Q}|}\sum_{q \in \mathcal{Q}}\mathbb{E}_{k}\left[\max_{1\leq j \leq k}r(\sigma_{q}(j)) \right].
\end{align}
In Eq~(\ref{eq:remove_theta}), by assuming $\theta \sim \text{Uni}(0, 1)$, we can obtain the following:
\begin{eqnarray}
  \mathbb{E}_{\theta}\left[\mathds{1}\left\{\max_{1\leq j \leq k}r(\sigma_{q}(j)) > \theta\right\}\right]=\max_{1\leq j \leq k}r(\sigma_{q}(j)).
\end{eqnarray}
Because we assume a uniform distribution for $k$ on $1 \leq k \leq K$ and $P_A(k)=1/K$,
we obtain the following:
\begin{align}
\notag
\text{(RHS of Eq.~(\ref{eq:remove_theta}))}
&= \frac{1}{|\mathcal{Q}|}\sum_{q \in \mathcal{Q}}\frac{1}{K}\sum_{k=1}^{K}\max_{1 \leq j \leq k}r(\sigma_{q}(j))\\
&= \frac{1}{|\mathcal{Q}|}\sum_{q \in \mathcal{Q}}\text{AxIoU@}K(q,\sigma).
\end{align}
That is, the mean AxIoU@$K$ can be considered as a marginalisation of the mean R@$K,\theta$ without any assumption for $\theta$ and with a weak assumption for $K$. 
The mean R@$K,\theta$ with a fixed value for each $K$ and $\theta$ evaluates certain aspects of systems' behaviour and thus requires to examine multiple settings of the parameters for evaluation.
We argue that AxIoU is a reasonable approach to avoiding the dependence on the $\theta$ threshold while considering the rank position of the best moment in a top-$K$ ranked list. 

\section{Requirements for Effectiveness Measures\label{section:axiom_definition}}
To evaluate the evaluation measures, we take an axiomatic approach.
We first set the following requirements for the design of our VMR measure based on the properties of R@$K,\theta$ in Section~\ref{section:conventional_measure}:
(1) It should ignore redundant moments in a ranked list,
(2) it should consider the IoU value between a ranked moment and the ground truth moment, and
(3) it should consider the rank position of relevant moments for evaluating top-$K$ retrieval effectiveness of a system.
We show that our Mean AxIoU satisfies all requirements while Mean R@$K,\theta$ satisfies only Requirement~(1).
To investigate VMR measures based on these requirements, we define two axioms for an effectiveness measure for VMR.

\begin{figure}[t!]
    \centering
    \includegraphics[clip,width=0.5\linewidth]{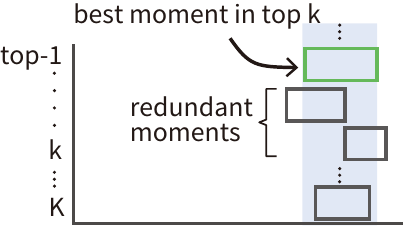}
    \caption{\textbf{INV-k} requires that a measure should be invariant to redundant moments which have smaller IoU and lower rank position than the best moment in the top-$k$ ($1 \leq k \leq K$) ranked list.}
    \label{fig:inv_k_visualisation}
\end{figure}

\paragraph{Invariance against Redundant Moments}
A measure should be unchanged to redundant moments in a ranked list.
We define this requirement as the following axiom.
\begin{axiom}[Invariance against Top-$k$ Non-Best Moment (\textbf{INV-k}).]
  \label{ax:inv-k}
  Suppose that two systems $\sigma$ and $\sigma'$ such that $\sigma'$ differs from $\sigma$ only for the $k$-th moment in the ranked lists for $q$.
  The measurement of $\sigma'$ must not change from that of $\sigma$ (i.e. $\mu(\mathcal{Q}, \sigma) = \mu(\mathcal{Q}, \sigma')$)  
  when the $k$-th moment in $\sigma'$ has a better IoU value than the $k$-th moment in $\sigma$ but is not the most relevant within the top $k$ of $\sigma'$.
\end{axiom}
Figure~\ref{fig:inv_k_visualisation} depicts the concept of \textbf{INV-k}.
R@$K,\theta$ satisfies this requirement because it utilises only the moment with the maximum IoU value in a ranked list (see Eq.~(\ref{eq:definition_recall})).
AxIoU can also handle the redundant moments by inheriting the property of R@$K,\theta$.
On the other hand, AP@$K,\theta$, which is a ranked retrieval measure widely adopted in computer vision~\cite{everingham2015pascal}, does not satisfy \textbf{INV-k}. 
Similarly, while an information retrieval measure for graded relevance such as DCG~\cite{jarvelin2002cumulated} would be a straightforward choice for evaluating ranked lists while avoiding the binarization by $\theta$,
it does not satisfy \textbf{INV-k} either. 
The formal definition of the axiom and proofs are given in our supplementary material.

\begin{figure}[t!]
    \centering
    \includegraphics[clip,width=0.98\linewidth]{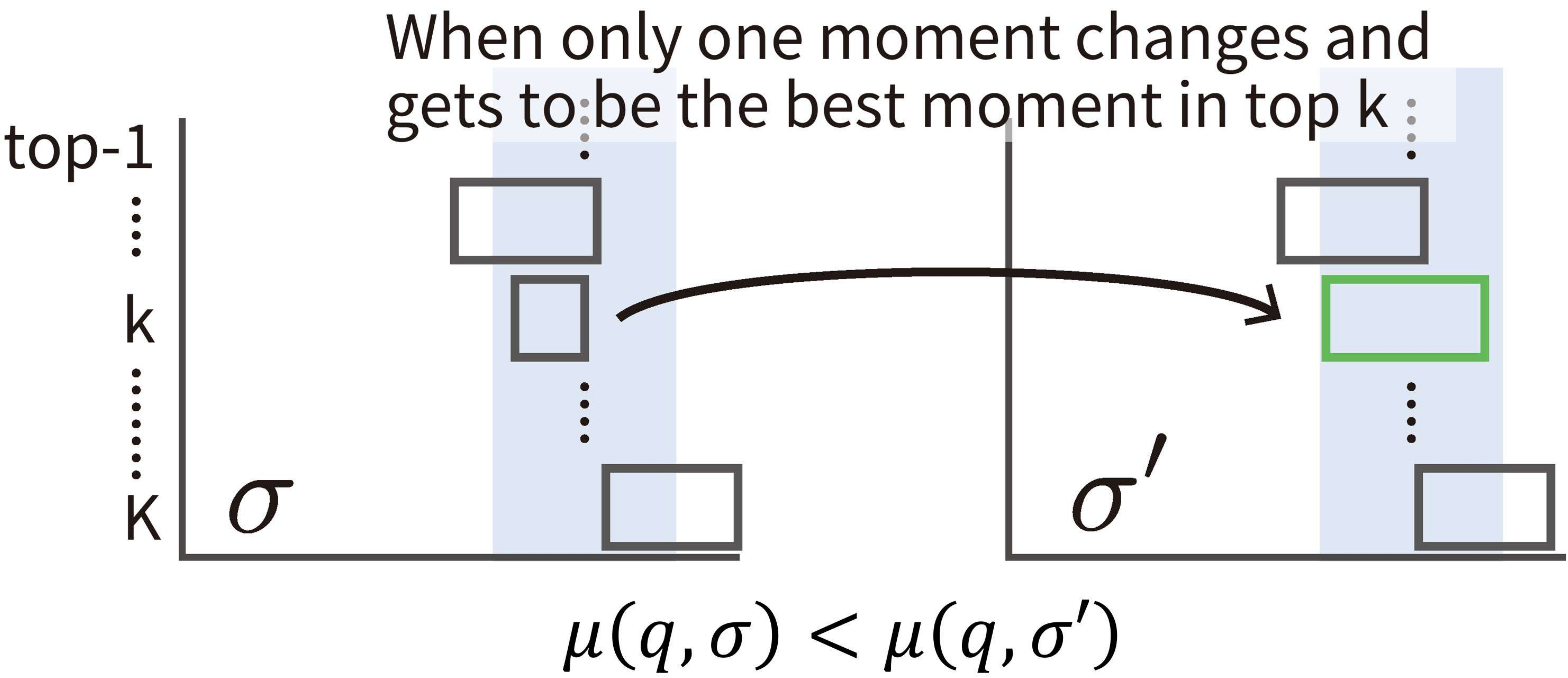}
    \caption{\textbf{MON-k} requires that a measure should be sensitive to the IoU value of the best moment in a top-$k$ ranked list.
    }
    \label{fig:mon_k_visualisation}
\end{figure}

\paragraph{Monotonicity with respect to the Best Moment}
The VMR measure score should monotonically increase with the maximum IoU value in a ranked list.
More specifically, we require that at any rank $k$, the measurement based on the top-$k$ moments of the SERP should monotonically increase with the maximum IoU observed within the top $k$.
This requirement can be defined through the following axiom.
\begin{axiom}[Strict Monotonicity for Top-$k$ Best Moment (\textbf{MON-k}).]
  \label{ax:mon-k}
  Suppose that two systems $\sigma$ and $\sigma'$ such that $\sigma'$ differs from $\sigma$ only for the $k$-th moment in the ranked lists for $q$.
  The measurement of $\sigma'$ strictly increases from that of $\sigma$ (i.e. $\mu(\mathcal{Q}, \sigma) < \mu(\mathcal{Q}, \sigma')$)  
  when the $k$-th moment in $\sigma'$ has a better IoU value than the $k$-th moment in $\sigma$ and is the most relevant within the top $k$ of $\sigma'$.
\end{axiom}
Figure~\ref{fig:mon_k_visualisation} depicts the concept of \textbf{MON-k}.
R@$K,\theta$ with a fixed parameter setting for $K$ and $\theta$ does not satisfy this requirement.
$\mu(\mathcal{Q},\sigma) <\mu(\mathcal{Q}, \sigma')$ is not guaranteed since R@$K,\theta$ binarizes the relevance using $\theta$.
By contrast, ranked retrieval measures for graded relevance, such as DCG@$K$ and our AxIoU@$K$, satisfies this property
because these consider the ranked position and IoU value of each moment in a ranked list.
The formal definition of the axiom and proofs are provided in the supplementary material.

\begin{figure*}[t!]
    \centering
    \includegraphics[clip,width=0.98\linewidth]{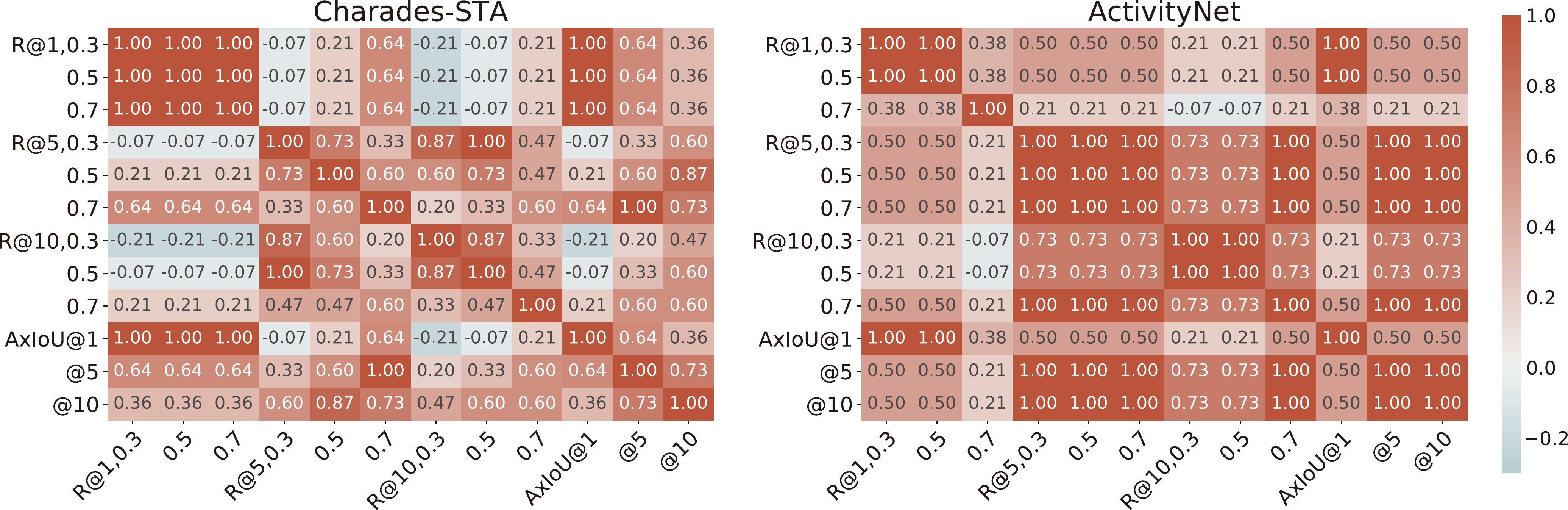}
    \caption{Agreement between two measures on Charades-STA (left) and ActivityNet (right).}
    \label{fig:agreement_matrix}
\end{figure*}

\section{Experiments}
While we analytically showed the properties of AxIoU@$K$ in terms of the axioms,
we also examine the measures empirically in this section.
We first investigate the agreement between the evaluation results based on the measures to confirm the compatibility of AxIoU@$K$ with R@$K,\theta$.
To examine the effect of $\theta$,
we also discuss the stability of the measures with respect to change in the test data.
Moreover, we demonstrate the advantages of AxIoU@$K$ as the criterion for model selection.

\subsection{Experimental Setup\label{exp:setup}}
\paragraph{Datasets}
Following the experimental settings of Otani \etal~\cite{otani2020challengesmr},
we utilise two popular datasets for our experiments, Charades-STA~\cite{gao2017tall} and ActivityNet~\cite{caba2015activitynet,krishna2017dense}.
Each dataset contains a set of manually annotated temporal regions for query-video pairs that indicate the relevant moment in a video as ground truth.
Charades-STA is built upon Charades~\cite{sigurdsson2016hollywood} and contains 9,848 videos, each of which is associated with multiple natural language sentences.
The number of test queries is 3,720. 
ActivityNet contains 19,209 YouTube videos. Each video is associated with the captions and their temporal locations.
The number of the test queries is 17,031.

\paragraph{Retrieval Systems for Evaluation}
In our experiment, we utilise multiple VMR systems to evaluate the measures;
for example, we create two rankings of the systems based on two measures, 
and then compute the similarity of the rankings (i.e. Kendall's $\tau\mbox{-}b$~\cite{agresti2010analysis}) as agreement between the two measures.
To examine each measure in a realistic setting, we employ real VMR systems trained on each dataset.
Throughout this paper,
we used three conventional methods, \textbf{Action-Aware Blind} (Blind)~\cite{otani2020challengesmr}, \textbf{SCDM}~\cite{yuan2019semantic} and
\textbf{2DTAN}~\cite{zhang2019learning}.
In addition, we include the variants of 2DTAN, \ie, (1) \textbf{2DTAN nonms}, a variant without Non-maximum suppression (NMS)~\cite{neubeck2006efficient}, (2) \textbf{2DTAN rand}, a variant with randomisation of video frames proposed by Otani \etal \cite{otani2020challengesmr}, and (3) \textbf{2DTAN rand+nonms}, a variant without NMS and with randomisation.

\begin{figure}[h!]
    \centering
    \includegraphics[clip,width=\linewidth]{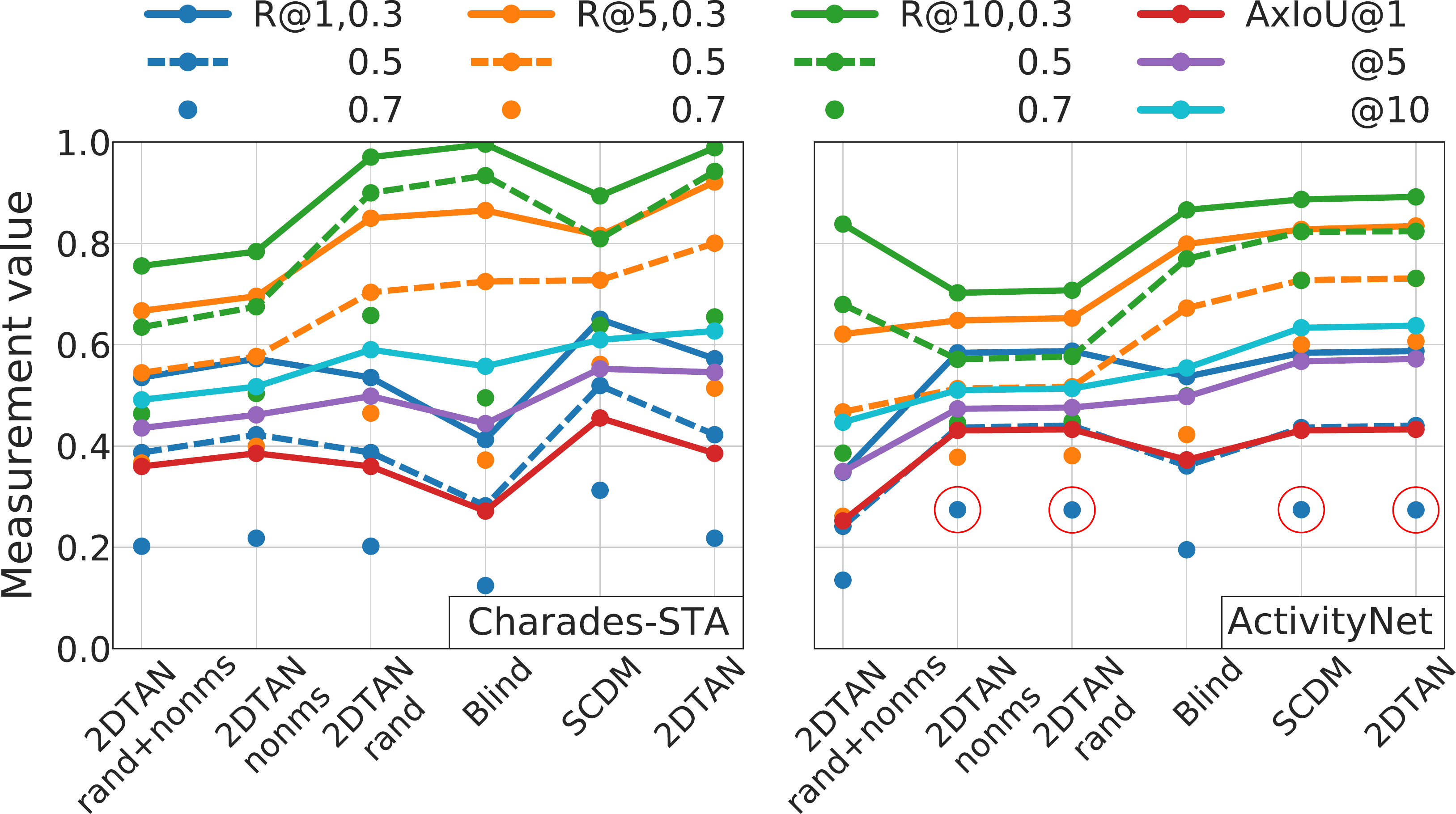}
    \caption{Effectiveness of each system on Charades-STA and ActivityNet datasets according to R@$K,\theta$ and AxIoU@$K$.}
    \label{fig:systems_measurement_values}
\end{figure}

Figure~\ref{fig:systems_measurement_values} compares the effectiveness of the above six systems
according to different measures on Charades-STA (left) and ActivityNet (right).
In each graph, the systems have been sorted by Mean R@$5,0.5$.
For Charades-STA, the R@$10,0.3$ score of Blind (green solid line), which is a video-agnostic baseline, is almost one;
as the Charades-STA dataset is a relatively easy dataset,
R@$10,0.3$ is a too relaxed measure even for Blind.
This result suggests that a inappropriate choice of $K$ and $\theta$ leads to uninformative evaluation results.

\subsection{Agreement between Measures\label{exp:agreement}}
Figure~\ref{fig:agreement_matrix} shows the agreement between each pair of measures among R@$K,\theta$ ($K=1,5,10, \theta=0.3,0.5,0.7$) and the AxIoU@$K$ ($K=1,5,10$) on Charades-STA and ActivityNet datasets, respectively.
To assess the agreement between two measures, we first rank the six systems using each measure.
We then compute Kendall's $\tau\mbox{-}b$~\cite{agresti2010analysis}, which considers the ties in a ranking.
Hereafter, we shall refer to $\tau\mbox{-}b$ simply as $\tau$.
A high $\tau$ value means that the rankings according to the two measures are similar~\cite{buckley2004retrieval}.

In the Charades-STA dataset, AxIoU@$10$ agrees well with all instances of R@$K,\theta$ ($0.36 \leq \tau \leq 0.87$).
The values of AxIoU@$K$ with different values of $K$ agree reasonably well with one another ($0.36 \leq \tau \leq 0.73$).
By contrast, different instances of R@$K,\theta$ can conflict with one another;
R@$5,0.7$ agrees well with R@$1,\theta$ ($\tau=0.64$) whereas R@$10,\theta$ does not agree with R@$1,\theta$ instances ($-0.21 \leq \tau \leq 0.21$). 
Probably, the main reason for this result is that R@$K,\theta$ is rank-insensitive.
On the other hand, AxIoU@$K$, which satisfies \textbf{MON-k}, aligns well with itself for different values of $K$.
Although R@$5,0.5$ and R@$5,0.7$, which are popular instances of R@$K,\theta$, agree relatively well with the other R@$K,\theta$ instances ($0.21 \leq \tau \leq 0.73$ for R@$5,0.5$ and $0.2 \leq \tau \leq 0.64$ for R@$5,0.7$),
the agreement between the two measures is $\tau=0.60$ despite the small difference in the setting of $\theta$;
remarkably, AxIoU@$10$ agrees with R@$5,0.5$ and R@$5,0.7$ with $\tau=0.87$ and $\tau=0.73$, respectively.

Because the ActivityNet dataset has a much larger number of test queries than that of the Charades-STA dataset,
most of the measures agree well with each other.
Nevertheless, R@$1,0.7$, which is a widely adopted instance of R@$K,\theta$, does not agree with the other instances ($-0.07 \leq \tau \leq 0.38$).
It is worth mentioning that R@$1,0.7$ is highly demanding (\ie requiring systems to return a highly relevant moment at rank 1).
Thus, R@$1,0.7$ lacks the sensitivity to distinguish between the systems.
As shown in Fig.~\ref{fig:systems_measurement_values} (red circles in the right side), the scores by R@$1,0.7$ are low for all six systems, and the scores for four out of six systems are all 0.274 (1,019/3,720).
AxIoU@$10$ achieves strong agreement ($\tau \geq 0.5$) with all instances of R@$K,\theta$ except R@$1,0.7$.

\begin{figure*}[t]
    \centering
    \includegraphics[clip,width=0.97\linewidth]{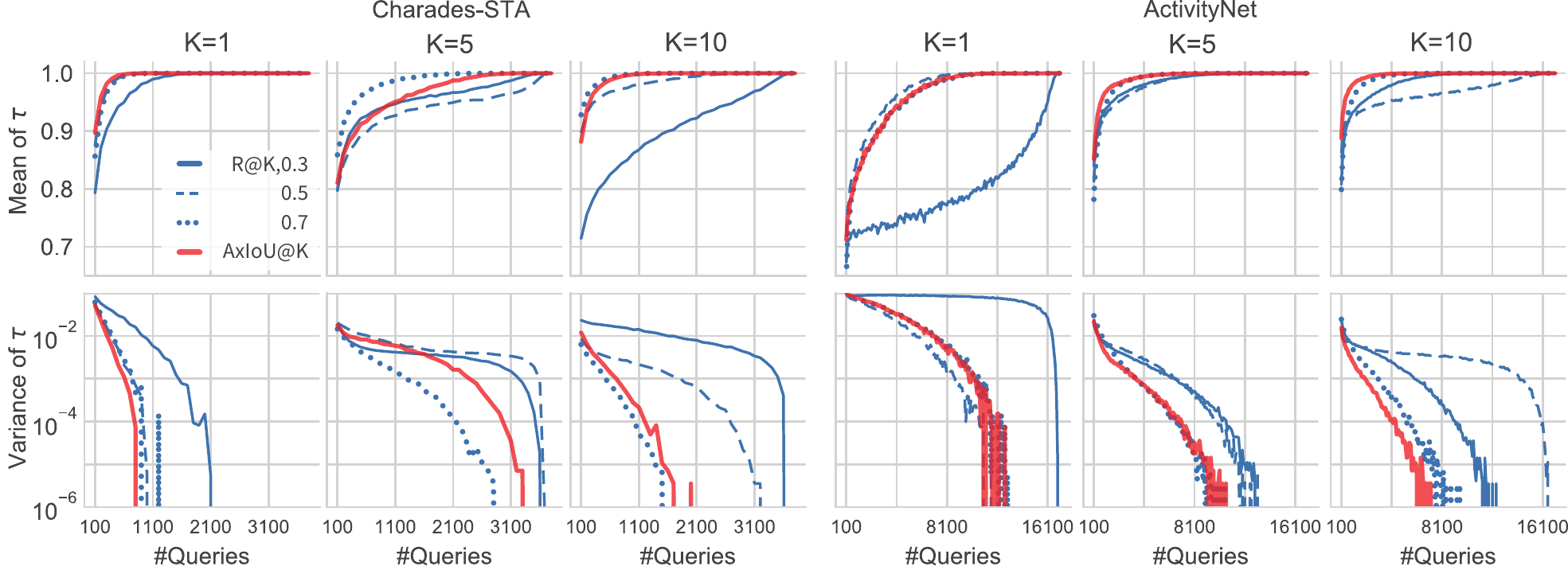}
    \caption{Effect of reducing the size of query subsets on means and variances of self-agreement for Charades-STA and ActivityNet.}
    \label{fig:self_agreement_curve_activitynet}
\end{figure*}

\subsection{Stability of Measures against the Choice of Evaluation Data\label{exp:stability}}
In this section, we investigate the stability of the measures,
\ie, the consistency of evaluation results based on a measure on different test datasets~\cite{buckley2004retrieval}.
The stability of an effective measure against different test datasets is one of the essential properties: If a measure is unstable, the conclusion drawn for a certain test dataset may not generalise well.
We evaluate each measure based on Kendall's $\tau\mbox{-}b$ between the system rankings based on two different subsets of a dataset as the \emph{self-agreement} of the measure.
To examine the stability with respect to the choice and size of test data
we investigate the self-agreement on conjoint query set pairs with different sizes.

Figure~\ref{fig:self_agreement_curve_activitynet}
visualises the effect of reducing the size of the query subsets on self-agreement.
We experimented with 5,000 trials for each query subset size.
The horizontal axes represent the size of each query subset.
The top graphs show the means of the self-agreement $\tau$'s;
the bottom graphs show the variances.

For Charades-STA in the first to third columns, it can be observed that,
for each $K$,
the R@$K,\theta$ instances with $\theta=0.3,0.5$ substantially underperform the other in terms of mean and variance of $\tau$;
on the other hand, the R@$K,0.7$ (red dotted line) instances are consistently stable.
The AxIoU@$K$ instances outperform most of the R@$K,\theta$ instances with the same value of $K$ whereas it performs relatively poorly with small query sets for $K=5$.
The most robust batch of measures for this dataset are R@$1,0.7$, R@$5,0.7$, R@$10,0.7$, AxIoU@$1$ and AxIoU@$10$.
Also for ActivityNet in the forth to fifth columns, the AxIoU@$K$ instances outperform most of the R@$K,\theta$;
the R@$K,0.7$ instances also perform well.

\subsection{Stability against Label Ambiguity\label{exp:uncertainty}}
In this section, we evaluate the measures in terms of the stability to label ambiguity (i.e. disagreement between human annotations).
We generate a testing sample based on a simple noise model by following steps;
(1) we consider each annotation in an original testing dataset as a low-noise sample and denote one by $(s^*, e^*) \in \mathbb{R}^2$ where $s^*$ and $e^*$ are the start and end points of a temporal boundary; 
(2) we draw a start point $s$ by a normal distribution with mean $s^*$ and variance $\beta^2$;
(3) we then draw a length $l$ by an exponential distribution with mean $e^* - s^*$; and
(4) we obtain the drawn sample $(s, s+l)$ as a noisy one.
For each testing sample in Charades-STA and ActivityNet,
we independently draw five samples from the noise model and then create a final testing annotation by taking medians of $s$ and $s+l$.
Here, it should be noted that the variance parameter $\beta^2$ can be considered as the quality (i.e. noise level) of five raters who annotate temporal boundaries to one sample.
We generate datasets with different noise levels by varying $\beta^2$ in $\{1, 2, 3, 4\}$.
The IoU between the mean of the median IoU values between the original and a drawn annotation for each noise level $\{1, 2, 3, 4\}$ is respectively $0.906$, $0.870$, $0.835$ and $0.802$ for Charades-STA, and $0.846$, $0.778$, $0.712$ and $0.650$ for ActivityNet;
note that, the noise levels are in a realistic range
as the previously reported IoU agreement between human annotations is around $0.725$ in Charades-STA~\cite{sigurdsson2017actions} and $0.641$ in ActivityNet~\cite{alwassel2018diagnosing}. 
We generate independent 100 testing datasets for each dataset and each noise level.
To evaluate the effect of label ambiguity for each measure, we compute the root mean squared error (RMSE) between the measurements based on the original dataset and 100 noisy datasets for each of six systems used in the above experiments.

\begin{figure*}[t]
    \centering
    \includegraphics[clip,width=0.99\linewidth]{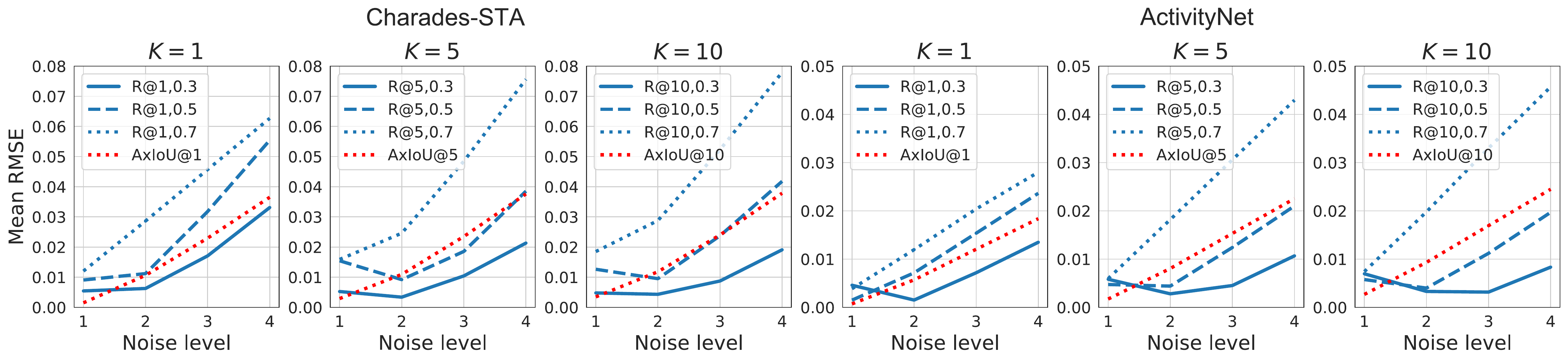}
    \caption{Effect of the ambiguity of annotated temporal regions.}
    \label{fig:stability_uncertainty}
\end{figure*}

Figure~\ref{fig:stability_uncertainty} shows the effect of the label noise on the evaluation based on the measures.
The x- and y-axes indicate the noise level and Mean RMSE for each measure.
In all datasets and all $K$, the AxIoU instances show lower errors than the R@$K,0.7$ instances but higher errors than R@$K,0.3$ instances in a wide range of noise levels.
In particular, the R@$K,0.7$ instances shows severely high errors.
This is because R@$K,\theta$ with large IoU threshold requires exactly localised moments thereby drastically changing evaluation results even with small perturbation in a ground truth.
Therefore, the use of a large $\theta$ assumes the low-noise condition of human annotations, which is difficult to ensure~\cite{hendricks17iccv,alwassel2018diagnosing,otani2020challengesmr}.
On the other hand, the instances of R@$K,\theta$ with $\theta=0.3,0.5$ show comparable or lower errors than the AxIoU instances because these R@$K,\theta$ instances ignore localisation quality. 

\subsection{Summary: Agreement and Stability}
We demonstrated the undesirable properties of R@$K,\theta$ from various aspects;
(1) the R@$K,0.3$ and R@$1,\theta$ instances (i.e. non-demanding measures) often disagree with other R@$K,\theta$ instances (Section~\ref{exp:agreement});
(2) the R@$K,0.3$ and R@$K,0.5$ instances are unstable to the change of the size of a dataset (Section~\ref{exp:stability}); and
(3) the R@$K,0.7$ instances are unstable to label ambiguity and potentially noisy (Section~\ref{exp:uncertainty}).
By contrast, our AxIoU measure reconciles the agreement and stability while reducing the hyper-parameter $\theta$, which is difficult to tune.
Moreover, it should be noted that
the cut-off parameter $K$ for AxIoU@$K$ is easier to handle than that of R@$K,\theta$
as it considers the ranking quality of top-$K$ ranked lists;
the agreement between the AxIoU instances (Section~\ref{exp:agreement}) is also an evidence for this.

\subsection{Model Selection\label{exp:model_selection}}
As discussed in Section \ref{exp:stability},
the stability with respect to the choice of test queries is vital for avoiding inconsistent evaluation on different dataset splits.
This is true also in the process of model selection;
when we select the best model based on a validation split and evaluate it on a test split,
the measurement on the validation split should be consistent with that on the test split.

This section investigates the effectiveness of AxIoU as the criterion for model selection.
To this end,
we first created 640 variants of the 2DTAN system (See Section~\ref{exp:setup})
by varying its hyper-parameters such as the learning rate and the threshold of NMS.
Then, based on each instance of R@$K,\theta$ as well as AxIoU (12 measures in total), 
we select the best model using the validation split.
Finally, we evaluate the above 12 models using R@$K,\theta$ on the test split.
As the R@$10,0.3$ scores saturate easily (see also Figure~\ref{fig:systems_measurement_values}),
we omitted it from the test measures for visibility of figures; thus, we utilise 8 test measures in total.
For each of the 8 test measures,
we compute the Z-scores of the 12 models
so that the average of the 12 scores equals zero.

\begin{figure}[t!]
    \centering
    \includegraphics[clip,width=0.92\linewidth]{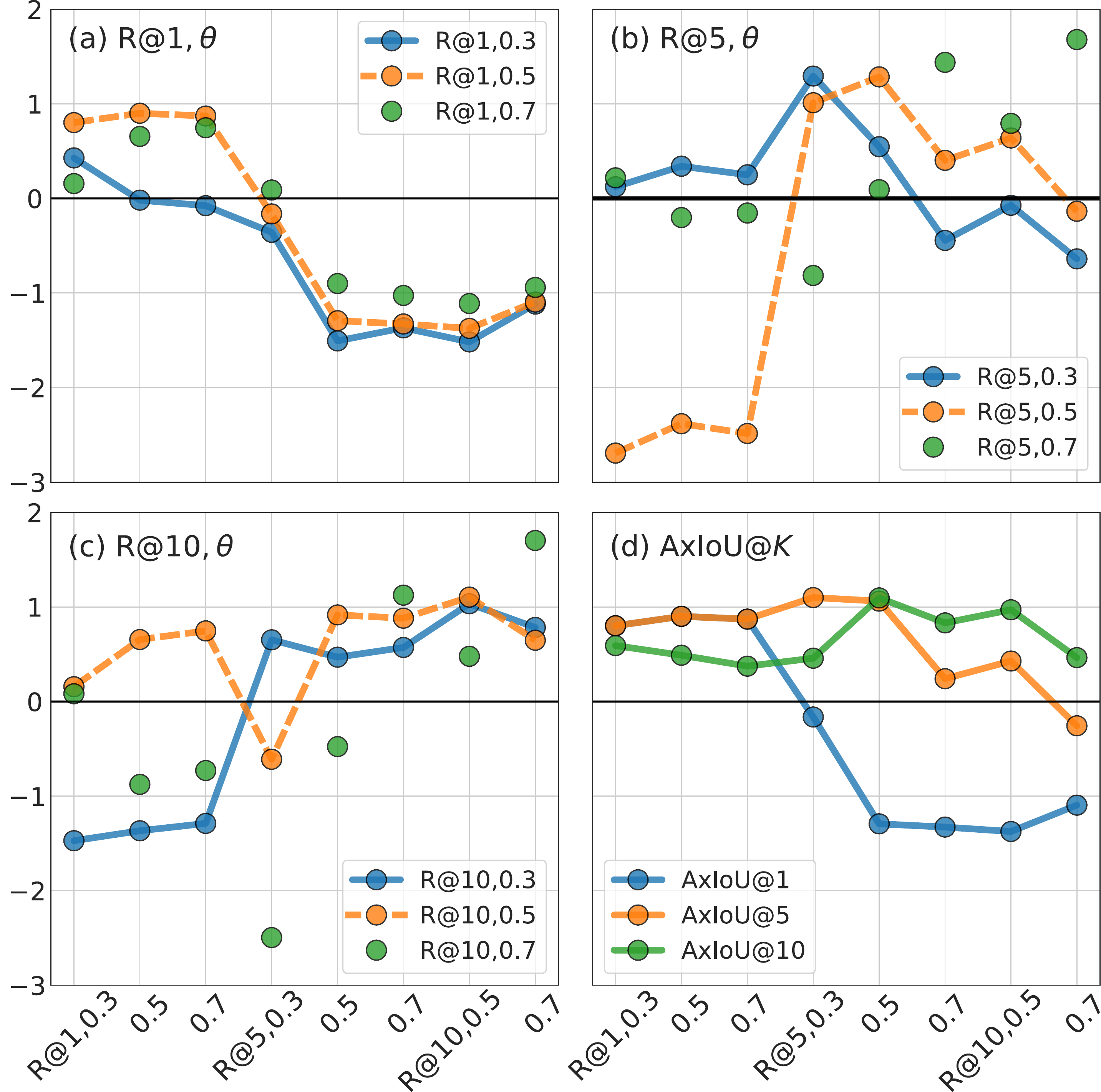}
    \caption{The effect of the validation measure for model selection on effectiveness on the test split.}
    \vspace{-0.3cm}
    \label{fig:model_selection}
\end{figure}

Figures~\ref{fig:model_selection} (a)--(d) show the results for Charades-STA.
The x-axis in each figure shows the test measure; each of the 12 lines represents a validation measure; the y-axis shows the Z-scores of ``all'' the 12 models for each test measure.
Since each line represents a single model selected by a particular validation measure, if the line is straight and horizontal, it would imply that the validation measure is useful for effective model selection.
R@$10,0.3$ (blue line in (c)) and R@$5,0.5$ (orange line in (b)) perform poorly as validation measures: when the models selected according to these measures are evaluated with R@$1,0.7$ on the test data, these systems are actually the worst among the 12 systems by far.
Similarly, R@$10,0.7$ (green in (c)),
R@$1,0.3$ (blue in (a)), and 
AxIoU@1 (blue in (d)) perform relatively poorly: for example,
when the model selected according to 
AxIoU@1 is evaluated with R@$10,0.5$ on the test data, this system is one of the worst performers among the twelve.
On the other hand, it can be observed that AxIoU@5, AxIoU@10 and some other R@$K, \theta$ instances such as R@$10,0.5$ (orange in (c)) perform well: that is, the models selected based on these measures generally perform well regardless of what the test measure is.
Only AxIoU@10 (green in (d)) could select a system that is above average in terms of all the test measures.

The above result is consistent to the insights obtained in Sections~\ref{exp:agreement}-\ref{exp:uncertainty}.
However, the disagreement and instability of each R@$K,\theta$ instance is severe for model selection
because we must use a single evaluation measure to determine the best model on a validation split.
As we cannot know the best setting of $K$ and $\theta$ in the validation phase,
AxIoU, which is an expectation of R@$K,\theta$ (Section~\ref{section:axiou_interpretation}), is a reasonable measure for model selection.

\section{Conclusion}
In this paper, we proposed an evaluation measure, AxIoU, for video moment retrieval.
AxIoU can offer consistent evaluation compared to R@$K,\theta$
without the threshold parameter $\theta$ in R@$K,\theta$, 
which is the main cause of the insensitivity of R@$K,\theta$.
We analytically examined the properties of AxIoU through an axiomatic approach
and empirically showed that AxIoU@$10$ can provide stable evaluation while maintaining the similarity to R@$K,\theta$ instances. 
We also demonstrated that AxIoU@$10$ is a reliable measure for model selection, even if the final test measures are R@$K,\theta$ instances.
As future work, we will explore a more sophisticated distribution for abandonment position $k$, $P_A(k)$~\cite{chapelle2009expected}.

\section{Acknowledgement}
This work was partly supported by JST CREST Grant No.~JPMJCR20D3, FOREST Grant No.~JPMJFR216O, and Academy of Finland project number 324346.

\appendix
\section{Proofs}
\setcounter{axiom}{0}

\subsection{Definition of Axioms}
We formally define the definition of each axiom and each measure in this supplementary material.
\begin{axiom}[Invariance against Top-$k$ Non-Best Moment (\textbf{INV-k}).]
  \label{ax:inv-k}
  For any query $q \in \mathcal{Q}$ and any rank position $k$ ($k>1$), and for all systems $\sigma$ and $\sigma'$ such that
  $\sigma'$ differs from $\sigma$ only for the $k$-th moment in the ranked lists for $q$,
  $\mu(\mathcal{Q}, \sigma) = \mu(\mathcal{Q}, \sigma')$ holds when the $k$-th moments satisfy the following conditions.
   \begin{condition}[Inequality of relevance scores]
    \label{cond:rel-score-ineq}
    The relevance scores of the $k$-th moment in $\sigma_q$ and $\sigma_q'$ satisfy   
    $r(\sigma_{q}(k)) < r(\sigma_{q}'(k))$.
  \end{condition}
  \begin{condition}[Non-maximum relevance score of the top-$k$ moment]
    \label{cond:non-max-rel-score}
    The $k$-th moment returned by system $\sigma$ is less relevant than that returned by system $\sigma'$. That is,    
    $r(\sigma_{q}'(k)) \leq \max_{1 \leq j < k}r(\sigma_{q}'(j))$.
  \end{condition}
\end{axiom}

\begin{axiom}[Strict Monotonicity for Top-$k$ Best Moment (\textbf{MON-k}).]
  \label{ax:mon-k}
  For any query $q \in \mathcal{Q}$ and any rank position $k$, and for all systems $\sigma$ and $\sigma'$ such that
  $\sigma'$ differs from $\sigma$ only for the $k$-th moment in the ranked lists for $q$,
  $\mu(\mathcal{Q}, \sigma) < \mu(\mathcal{Q}, \sigma')$ holds whenever the $k$-th moment satisfies Condition~\ref{cond:rel-score-ineq} and the following condition. 
  \begin{condition}[Maximum relevance score of the top-$k$ moment]
    \label{cond:max-rel-score}
    The $k$-th moment returned by $\sigma'$ is the most relevant within the top $k$. 
    That is,
    $r(\sigma_{q}'(k)) > \max_{1 \leq k' < k}r(\sigma_{q}'(k'))$ if $k>1$.    
  \end{condition}
\end{axiom}

Note that, Condition~\ref{cond:max-rel-score} is necessary to avoid the contradiction between \textbf{INV-k} and \textbf{MON-k}.

\subsection{Properties of R@$K,\theta$\label{section:recall_proofs}}
\text{Mean R}@$K,\theta$, is defined as the ratio of queries for which a system successfully retrieves at least one relevant moment with a sufficient IoU with respect to threshold $\theta$~\cite{gao2017tall}.
\begin{align}
  \notag
  &\text{Mean R}@K,\theta(\mathcal{Q}, \sigma) \\
  &= \frac{1}{|\mathcal{Q}|}\sum_{q \in \mathcal{Q}}\mathds{1}\left\{\sum_{k=1}^{K}\mathds{1}\left\{r(\sigma_{q}(k)) > \theta\right\} > 0\right\}.
\end{align}

\begin{property}
  \label{property:recall-mon-k}
  \text{Mean R}@$K,\theta$ does not satisfy \textbf{MON-k} (Axiom~\ref{ax:mon-k}).
\end{property}

\begin{proof}
  For two systems $\sigma$ and $\sigma'$ such that $\sigma'$ differs from $\sigma$ only for $k$-th moment in the ranked list for $q$,
  the difference of the measurements can be expressed as follows:  
  \begin{align}
    \notag
    &\text{Mean R}@K,\theta(\mathcal{Q}, \sigma) - \text{Mean R}@K,\theta(\mathcal{Q}, \sigma') \\\notag
    & = \frac{1}{|\mathcal{Q}|}\mathds{1}\left\{\mathds{1}\left\{r(\sigma(q)_{k}) > \theta\right\}+C > 0\right\} \\\label{eq:recall-diff}
    & - \frac{1}{|\mathcal{Q}|}\mathds{1}\left\{\mathds{1}\left\{r(\sigma'(q)_{k}) > \theta\right\}+C > 0\right\},
  \end{align}
  where $C=\sum_{1\leq j \leq K \land j \neq k}^{K}\mathds{1}\left\{r(\sigma(q)_{j}) > \theta\right\}$.
  Here, when $r(\sigma_{q}'(k)) \leq \theta$ holds,
  it also holds that $r(\sigma_{q}(k)) \leq \theta$ by utilising Condition~\ref{cond:rel-score-ineq}.
  Then, the $k$-th moments do not contribute to the measurements, $\mathds{1}\left\{r(\sigma_{q}(k)) > \theta\right\}=\mathds{1}\left\{r(\sigma_{q}'(k)) > \theta\right\}=0$.
  Here, because $\theta \geq r(\sigma_{q}'(k)) \geq \max_{1 \leq j < k}r(\sigma_{q}'(j))$ holds by Condition~\ref{cond:max-rel-score},
  there is no moment that has a sufficient relevance score in the ranked lists $\sigma_q$ and $\sigma_q'$ and $C=0$ holds.
  Therefore, combining these and Eq.~(\ref{eq:recall-diff}),
  when $r(\sigma_{q}'(k)) \leq \theta$, 
  we obtain $\text{Mean R}@K,\theta(Q, \sigma)=\text{Mean R}@K,\theta(Q, \sigma')$, which proves our proposition. 
\end{proof}
This problem results from the thresholding of temporal IoUs in the measure.
This leads to the information loss of the retrieval effectiveness by binarizing the relevance score of moments
and thus to the insensitivity of the measure.
Property~\ref{property:recall-mon-k} suggests that
the measure may ignore the improvement of systems when utilising a large value of $\theta$.

Remarkably, R@$K,\theta$ obviously does not satisfy \textbf{MON-k} even with assuming $r(\sigma_q(k)) > \theta$ in the case of $K > 1$;
when $r(\sigma_q(j)) > \theta$ holds for any rank position $j$ $(1 \leq j \leq K \land j \neq k)$,
$C \geq 1$ in Eq.~(\ref{eq:recall-diff}) holds, and thus $\text{Mean R}@K,\theta(\mathcal{Q}, \sigma) - \text{Mean R}@K,\theta(\mathcal{Q}, \sigma')= (1/|\mathcal{Q}|)(1 - 1) = 0$ holds.
Therefore, setting a small value of $\theta$,
it also leads to information loss.
%% This problem also results from the thresholding.
%% It is worth mentioning that
%% a small value of $\theta$ also leads the insensitivity
%% because moments with a wide range of relevance magnitude are considered as relevant ones.

\begin{property}
  \label{property:recall-inv-k}
  \text{Mean R}@$K,\theta$ satisfies \textbf{INV-k} (Axiom~\ref{ax:inv-k}).
\end{property}

\begin{proof}  
  Because we may assume that Condition~\ref{cond:non-max-rel-score} holds,
  when $r(\sigma_q'(k)) > \theta$, there is at least one moment in a position $j$ that satisfies $r(\sigma_{q}'(j)) \geq r(\sigma_{q}'(k)) > \theta$,
  and thus, $C \geq 1$ holds in Eq.~(\ref{eq:recall-diff}).
  When $r(\sigma_{q}'(k)) \leq \theta$, the $k$-th moment does not contribute to the measurement, and $\mathds{1}\left\{r(\sigma_{q}'(k)) > \theta\right\}=0$ holds.
  Therefore, by utilising Condition~\ref{cond:rel-score-ineq}, $r(\sigma_{q}(k)) \leq r(\sigma_{q}'(k)) \leq \theta$, we have,
  \begin{align}
    \notag
    &\text{Mean R}@K,\theta(\mathcal{Q}, \sigma) - \text{Mean R}@K,\theta(\mathcal{Q}, \sigma') \\
    & = \frac{1}{|\mathcal{Q}|}\mathds{1}\left\{C > 0\right\} - \frac{1}{|\mathcal{Q}|}\mathds{1}\left\{C > 0\right\} = 0,
  \end{align}
  regardless of $r(\sigma'(q)_{k}) > \theta$ or $r(\sigma'(q)_{k}) \leq \theta$.  
  Thus, we obtain $\text{Mean R}@K,\theta(Q, \sigma)=\text{Mean R}@K,\theta(Q, \sigma')$, which proves our proposition.
\end{proof}

This result suggests that the thresholding and indicator function in R@$K,\theta$ play a vital role in ensuring
invariance against the redundant moments in the lower rank positions.
Although these mechanisms are indispensable as the invariance is required under the problem settings of VMR,
they are the main causes of information loss (See Property~\ref{property:recall-mon-k}).

\subsection{Properties of AP Measures\label{section:map_proofs}}
Using the average precision (AP) measure is one approach to consider the rank of relevant moments~\cite{manning2008introduction}.
AP and Mean AP (a.k.a. mAP) can be expressed as follows:
\begin{align}
  &\text{AP}@K,\theta(q, \sigma) \coloneqq \frac{1}{K}\sum_{k=1}^{K}\frac{1}{k}\sum_{j=1}^{k}\mathds{1}\{r(\sigma_q(j) > \theta)\}. \\\notag
  &\text{Mean AP}@K,\theta(\mathcal{Q}, \sigma) \\&\coloneqq \frac{1}{|\mathcal{Q}|}\sum_{q \in \mathcal{Q}}\frac{1}{K}\sum_{k=1}^{K}\frac{1}{k}\sum_{j=1}^{k}\mathds{1}\{r(\sigma_q(j) > \theta)\}.
\end{align}
As the AP measure is for binary relevance grades,
it also requires a thresholding process for IoU values.

\begin{property}
  \label{property:map-inv-k}
  Mean AP@$K,\theta$ does not satisfy \textbf{INV-k} (Axiom~\ref{ax:inv-k}).
\end{property}

\begin{proof}
  For two systems $\sigma$ and $\sigma'$ such that $\sigma'$ differs from $\sigma$ only for $k'$-th moment in the ranked list for $q$,
  the difference of the measurements can be expressed as follows:  
  \begin{align}
    \notag
    &\text{Mean AP}@K,\theta(\mathcal{Q}, \sigma') - \text{Mean AP}@K,\theta(\mathcal{Q}, \sigma) \\\notag
     & = \frac{1}{|\mathcal{Q}|K}\sum_{k=1}^{K}\frac{1}{k}\sum_{j=1}^{k}\left(\mathds{1}\{r(\sigma_q'(j) > \theta)\} - \mathds{1}\{r(\sigma_q(j) > \theta)\}\right) \\\label{eq:ap_diff}
     & = \frac{1}{|\mathcal{Q}|K}\sum_{k=1}^{K}\frac{1}{k}\left(\mathds{1}\{r(\sigma_q'(k') > \theta)\} - \mathds{1}\{r(\sigma_q(k') > \theta)\}\right).
  \end{align}
  To derive the second equality, we assume that the top-$(k'-1)$ ranked lists of $\sigma_q$ and $\sigma_q'$ are identical, and the partial ranked lists from the $(k'+1)$-th position are also identical.
  When $r(\sigma_q'(k')) > \theta \geq r(\sigma_q(k'))$ holds, we have the following: $\mathds{1}\{r(\sigma_q'(k') > \theta)\} - \mathds{1}\{r(\sigma_q(k') > \theta)\}=1-0=1$.
  Therefore, we can obtain the following:
   \begin{align}
    \notag
    &\text{Mean AP}@K,\theta(\mathcal{Q}, \sigma') - \text{Mean AP}@K,\theta(\mathcal{Q}, \sigma) \\\notag
     & = \frac{1}{|\mathcal{Q}|K}\sum_{k=1}^{K}\frac{1}{k} > 0 \\\notag
     &\Longleftrightarrow \text{Mean AP}@K,\theta(\mathcal{Q}, \sigma') > \text{Mean AP}@K,\theta(\mathcal{Q}, \sigma).
  \end{align}
\end{proof}
AP cannot handle the redundant moments in a ranked list because each top-$K$ ranked relevant moment contributes to the measurement as an equally relevant one; in other words, AP is concerned with the number of the relevant moments in a ranked list.
It suggests that a system without NMS can unfairly take an advantage in the evaluation based on AP.

\begin{property}
  \label{property:map-mon-k}
  Mean AP@$K,\theta$ does not satisfy \textbf{MON-k} (Axiom~\ref{ax:inv-k}).
\end{property}

\begin{proof}
  In Eq.~(\ref{eq:ap_diff}),
  when $\theta \geq r(\sigma_q'(k')) > r(\sigma_q(k'))$ and Condition~\ref{cond:max-rel-score} hold,
  $\mathds{1}\{r(\sigma_q'(k') > \theta)\} - \mathds{1}\{r(\sigma_q(k') > \theta)\}=0-0=0$.
  Therefore, we obtain the following:
   \begin{align}
    \notag
    &\text{Mean AP}@K,\theta(\mathcal{Q}, \sigma') - \text{Mean AP}@K,\theta(\mathcal{Q}, \sigma) \\\notag
     & = 0\\\notag
     &\Longleftrightarrow \text{Mean AP}@K,\theta(\mathcal{Q}, \sigma') = \text{Mean AP}@K,\theta(\mathcal{Q}, \sigma).
  \end{align}
\end{proof}
Although AP is rank-sensitive,
it has the threshold $\theta$ as in R@$K,\theta$ and can ignore the improvement of IoU values of relevant moments.

\subsection{Properties of DCG-type Measures\label{section:dcg_proofs}}
The na\"{i}ve approach to remove thresholding parameter $\theta$ while considering the rank positions of relevant moments
is to utilise the measures for multiple relevance grades, such as normalised discounted cumulative gain (nDCG)~\cite{jarvelin2002cumulated}
because an IoU value can be considered as a continuous relevance score.
A DCG-type measure can be expressed as follows:
\begin{align}
  \text{DCG}@K(q, \sigma) &\coloneqq \sum_{k=1}^{K}\frac{g(r(\sigma(q)_i))}{d(k)}. \\
  \text{Mean DCG}@K(\mathcal{Q}, \sigma) &\coloneqq \frac{1}{|\mathcal{Q}|}\sum_{q \in \mathcal{Q}}\sum_{k=1}^{K}\frac{g(r(\sigma(q)_i))}{d(k)},
\end{align}
where $g(r) \geq 0$ and $d(k)>0$ denotes the gain and discounting functions
that are strictly monotonically increasing with respect to the relevance score $r$ and rank position $k$, respectively.
However, any DCG-type measure obviously does not satisfy \textbf{INV-k} (Axiom~\ref{ax:inv-k}).
\begin{property}
  \label{property:dcg-inv-k}
  Mean DCG@$K$ does not satisfy \textbf{INV-k} (Axiom~\ref{ax:inv-k}).
\end{property}

\begin{proof}
  For any query $q \in \mathcal{Q}$ and any rank position $k$, and for all systems $\sigma$ and $\sigma'$ such that
  $\sigma'$ differs from $\sigma$ for only the $k$-th moment in the ranked lists for $q$,  
  we can obtain the following equation.
  \begin{align}
  \notag
    &\text{Mean DCG}@K(\mathcal{Q}, \sigma') - \text{Mean DCG}@K(\mathcal{Q}, \sigma) \\  \notag
    &= \frac{1}{|\mathcal{Q}|}\sum_{q \in \mathcal{Q}}\sum_{j=1}^{K}\frac{g(r(\sigma_{q}'(j)))}{d(j)} - \frac{1}{|\mathcal{Q}|}\sum_{q \in \mathcal{Q}}\sum_{j=1}^{K}\frac{g(r(\sigma_{q}(j)))}{d(j)} \\ \label{eq:ndcg_diff}
    &= \frac{1}{|\mathcal{Q}|}\frac{g(r(\sigma_{q}'(k)))-g(r(\sigma_{q}(k)))}{d(k)}.
  \end{align}
  Based on Condition~\ref{cond:rel-score-ineq}, $r(\sigma_{q}(k)) < r(\sigma_{q}'(k))$ and the strict monotonicity of the gain function $g(\cdot)$,
  we have the following: $\text{Mean DCG}@K(\sigma') - \text{Mean DCG}@K(\sigma) > 0$, which completes the proof. 
\end{proof}

Because DCG@$K$ is defined as the sum of element-wise discounted gain values of moments in a ranked list,
it cannot handle the redundant moments in a ranked list appropriately.
Moreover, it is difficult to utilise DCG-type measures with normalisation (nDCG) for VMR
as the definition of the ideal list is not trivial owing to \textbf{INV-k};
thus, DCG@$K$ is under-normalised and can take a large value for a single query.

\begin{property}
  \label{property:dcg-mon-k}
  Mean DCG@$K$ satisfies \textbf{MON-k} (Axiom~\ref{ax:inv-k}).
\end{property}

\begin{proof}
  In Eq.~(\ref{eq:ndcg_diff}),
  by the strict monotonicity of the gain function $g(\cdot)$ and non-negativity of the discount function $d(\cdot)$,
  $g(r(\sigma_{q}'(k)))-g(r(\sigma_{q}(k)))>0$, and thus, $\text{Mean DCG}@K(\sigma') - \text{Mean DCG}@K(\sigma) > 0$ always holds, which completes the proof.
\end{proof}
DCG@$K$ is thresholding-free and rank-sensitive, and thereby satisfies \textbf{MON-k};
however, due to these properties, it does not satisfies \textbf{INV-k}.

By considering the properties of R@$K,\theta$ and DCG@$K$,
it is challenging for conventional information retrieval measures to satisfy \textbf{INV-k} and \textbf{MON-k} simultaneously.

\subsection{Properties of AxIoU Measure\label{section:axiou_proofs}}
In this section, we demonstrate that AxIoU is thresholding-free and rank-sensitive while satisfying both \textbf{INV-k} and \textbf{MON-k}.   
AxIoU can be expressed as follows:
\begin{align}
  &\text{AxIoU}@K(q, \sigma) \coloneqq \frac{1}{K}\sum_{k=1}^{K}\max_{1 \leq j \leq k}r(\sigma_{q}(j)), \\
  &\text{Mean AxIoU}@K(\mathcal{Q}, \sigma) \coloneqq \frac{1}{|\mathcal{Q}|}\sum_{q \in \mathcal{Q}}\frac{1}{K}\sum_{k=1}^{K}\max_{1 \leq j \leq k}r(\sigma_{q}(j)).
\end{align}

\begin{property}
  \label{property:axiou-inv-k}
  Mean AxIoU@$K$ satisfies \textbf{INV-k} (Axiom~\ref{ax:inv-k}).
\end{property}

\begin{proof}
  For two systems $\sigma$ and $\sigma'$ such that $\sigma'$ differs from $\sigma$ only for the $k'$-th moment in the ranked list for $q$,
  \begin{align}
    \notag
    &\text{Mean AxIoU}@K(\mathcal{Q}, \sigma') - \text{Mean AxIoU}@K(\mathcal{Q}, \sigma)\\\label{eq:axiou_topk_diff_inv}
    = &\frac{1}{|\mathcal{Q}|K}\sum_{k=k'}^{K}\left(\max_{1 \leq j \leq k}r(\sigma_{q}'(j)) - \max_{1 \leq j \leq k}r(\sigma_{q}(j))\right)
  \end{align}
  By utilising $r(\sigma_{q}'(k')) \leq \max_{1 \leq j < k'}r(j)$ and $r(\sigma_{q}(k')) < r(\sigma_{q}'(k'))$,
  it holds that $r(\sigma_{q}(k')) \leq \max_{1 \leq j < k'}r(\sigma_{q}(j))$ because the top-$(k'-1)$ lists of $\sigma$ and $\sigma'$ are identical.
  Therefore, $\max_{1 \leq j \leq k'}r(\sigma_{q}'(j))=\max_{1 \leq j \leq k'}r(\sigma_{q}(j))$ holds.
  In addition, because the partial ranked lists of $\sigma$ and $\sigma'$ from the $(k'+1)$-th position are identical,
  we have $\max_{1 \leq j \leq k}r(\sigma_{q}'(j))=\max_{1 \leq j \leq k}r(\sigma_{q}(j))$ for any position $k (1 \leq k \leq K)$.
  Therefore, we have the following: 
  \begin{align}
  \notag
    &\sum_{k=k'}^{K}\left(\max_{1 \leq j \leq k}r(\sigma_{q}'(j)) - \max_{1 \leq j \leq k}r(\sigma_{q}(j))\right) = 0 \\\notag
    &\Longleftrightarrow \text{Mean AxIoU}@K(\mathcal{Q}, \sigma') = \text{Mean AxIoU}@K(\mathcal{Q}, \sigma).
  \end{align}
\end{proof}
AxIoU determines the contribution of the relevant moments in a ranked list by comparing the relevance of these moments.
Therefore, it can handle the redundant moments without any binarisation and thresholding processes.

\begin{property}
  \label{property:axiou-mon-k}
  Mean AxIoU@$K$ satisfies \textbf{MON-k} (Axiom~\ref{ax:mon-k}).
\end{property}

\begin{proof}
  For two systems $\sigma$ and $\sigma'$ such that $\sigma'$ differs from $\sigma$ only for $k'$-th moment in the ranked list for $q$,
  the evaluation measures can be expressed as follows:
  \begin{align}
    \notag
    &\text{Mean AxIoU}@K(\mathcal{Q}, \sigma') - \text{Mean AxIoU}@K(\mathcal{Q}, \sigma)\\\notag
    &= \frac{1}{|\mathcal{Q}|K}\sum_{k=k'}^{K}\left(\max_{1 \leq j \leq k}r(\sigma_{q}'(j)) - \max_{1 \leq j \leq k}r(\sigma_{q}(j))\right)\\\notag
    &= \frac{1}{|\mathcal{Q}|K}\Bigg(r(\sigma_{q}'(k'))-\max_{1 \leq j \leq k'}r(\sigma_{q}(j)) \\\label{eq:axiou_topk_diff}
    &+ \sum_{k=k'+1}^{K}\left(\max_{1 \leq j \leq k}r(\sigma_{q}'(j)) - \max_{1 \leq j \leq k}r(\sigma_{q}(j))\right)\Bigg)
  \end{align}
  In the second equality, we utilised $r(\sigma_{q}'(k')) = \max_{1 \leq j \leq k'}r(\sigma_{q}'(j))$ to derive the first term in the right hand side.
  By utilising $r(\sigma_{q}'(k')) > \max_{1 \leq j < k'}r(\sigma_{q}'(j))$ and $r(\sigma_{q}'(k')) > r(\sigma_{q}(k'))$,
  $r(\sigma_{q}'(k'))-\max_{1 \leq j \leq k'}r(\sigma_{q}(j)) > 0$ holds in the right hand side of the second equality.
  For the second term, because we may assume that the partial ranked lists of $\sigma$ and $\sigma'$ from the $(k'+1)$-th position are identical,
  $\max_{1 \leq j \leq k}r(\sigma_{q}'(j)) \geq \max_{1 \leq j \leq k}r(\sigma_{q}(j))$ holds for any position $k$  $(k' < k \leq K)$.
  Thus, the following inequality holds.
  \begin{align}
  \notag
    &r(\sigma_{q}'(k'))-\max_{1 \leq j \leq k'}r(\sigma_{q}(j)) > 0 \\\notag
    &\land  \sum_{k=k'+1}^{K}\left(\max_{1 \leq j \leq k}r(\sigma_{q}'(j)) - \max_{1 \leq j \leq k}r(\sigma_{q}(j))\right) \geq 0 \\\notag
    &\Longleftrightarrow \text{Mean AxIoU}@K(\mathcal{Q}, \sigma') > \text{Mean AxIoU}@K(Q, \sigma),
  \end{align}
  which completes the proof.
\end{proof}

As a summary, AxIoU reflects the rank positions of the relevant moments in a ranked list as in AP 
and considers IoU values as in DCG, while it can handle the redundant moments as in R@$K,\theta$.

\section{Analysis of Number of Tied Results}
\begin{figure}[h]
    \centering
    \includegraphics[clip,width=\linewidth]{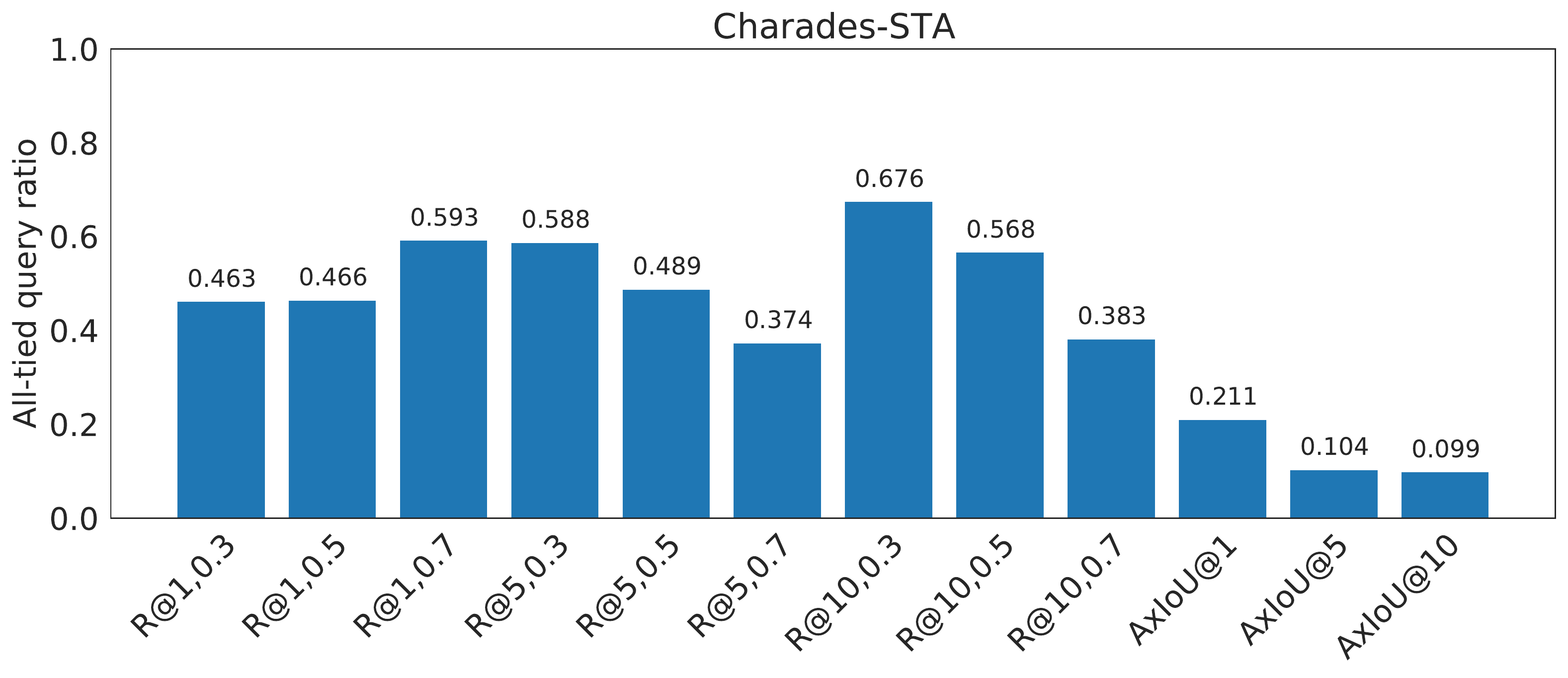}
    \includegraphics[clip,width=\linewidth]{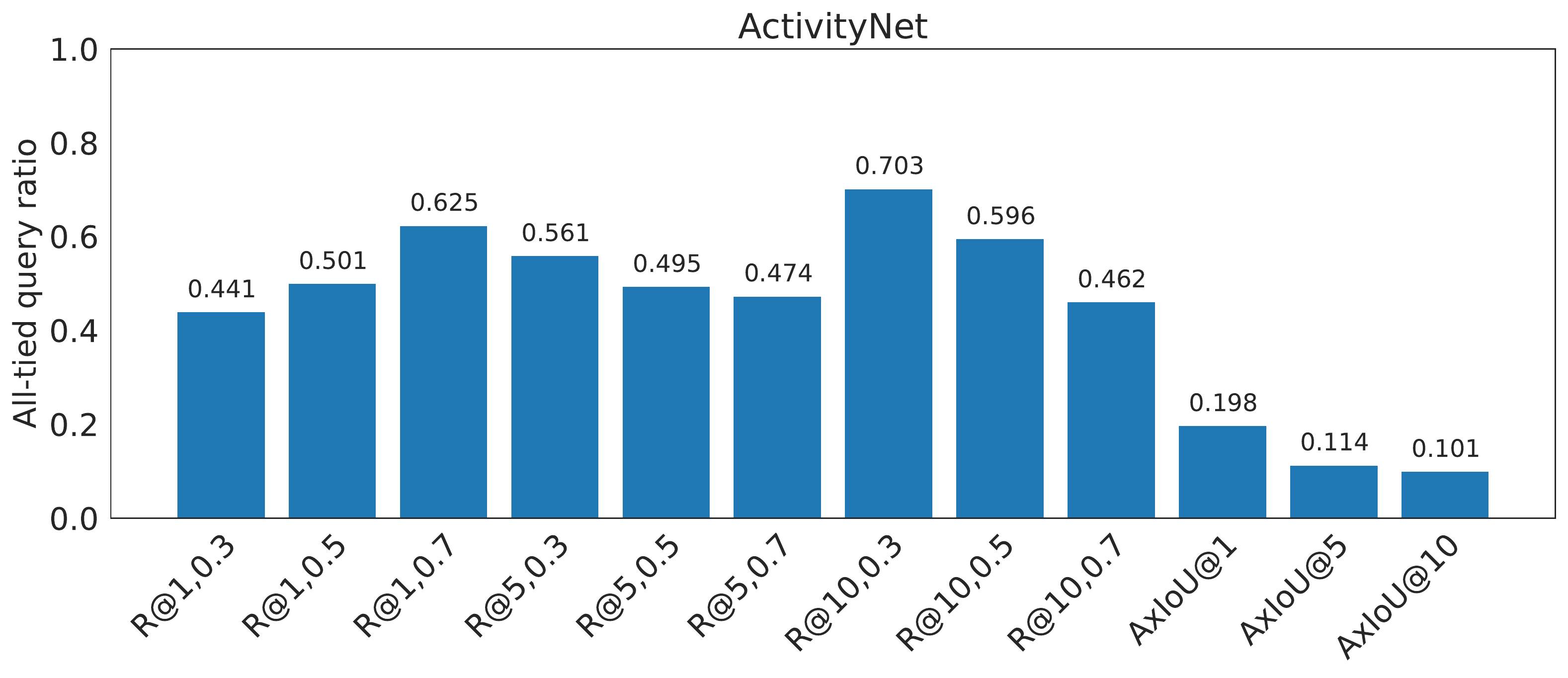}
    \caption{All-tied query ratio of each measure.
    }
    \label{fig:all_tie}
\end{figure}

To demonstrate the behaviours of R@$K,\theta$ and AxIoU@$K$,
we investigate the number of queries for which the 6 systems have the exactly same score for each VMR measure;
we define the ratio of such queries in all test queries as \emph{all-tied query ratio} of a measure.
Figure~\ref{fig:all_tie} shows the all-tied query ratio of each measure on Charades-STA and ActivityNet, respectively.
From Figure~\ref{fig:all_tie},
the R@$K,\theta$ instances with relaxed or demanding settings, such as R@$10,0.3$, R@$1,0.7$, show higher all-tied query ratios than the other instances.
It indicates that these measures cannot distil any information from the evaluation results based on a large number of queries. 
R@$5,0.7$ performs well in both Charades-STA and ActivityNet.
On the other hand, the AxIoU@$K$ instances show substantially lower all-tied query ratios for $K=1,5,10$.
It is remarkable that, with a larger $K$, AxIoU@$K$ performs well whereas R@$K,\theta$ with $\theta=0.3,0.5$ becomes worse.
Probably, it is because AxIoU@$K$ can leverage the information of the lower positions in ranked lists owing to its rank-sensitivity,
whereas R@$K,\theta$, which is a set retrieval measure, becomes insensitive when with a large $K$ and requires a large $\theta$ to detect the difference of systems.
This suggests that the setting of $\theta$ is rather difficult when $K$ is large such as in the TVR dataset~\cite{lei2020tvr};
an extremely large $\theta$ may be required although it can make difficult queries uninformative.

\begin{figure}[t]
    \centering
    \includegraphics[width=\linewidth]{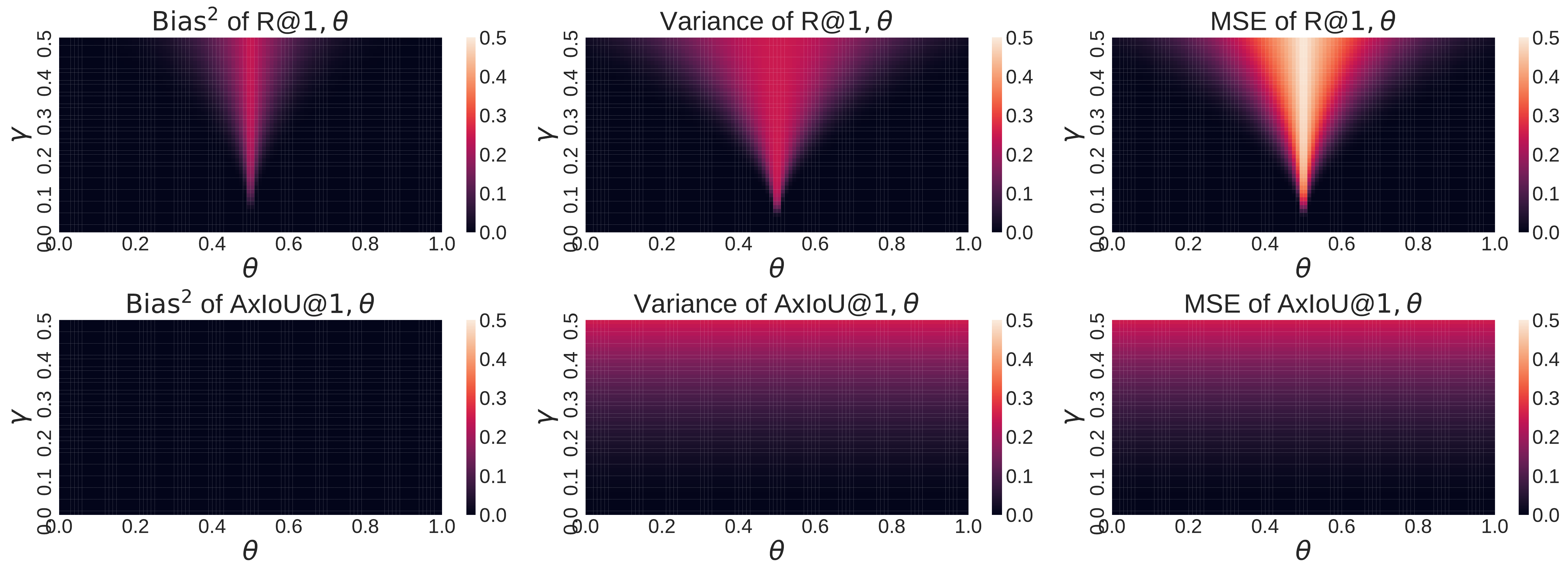}
    \caption{Effect of $\theta$ and $\gamma$ on estimation errors.}
    \label{fig:variance-example}
    \vspace{-0.5cm}
\end{figure}

\section{On the Stability to Label Ambiguity}
%The formal discussion on the stability to label ambiguity may be overwhelming for this paper as it requires analyses on the statistical property under a noise model of human assessments.
To show the stability to label ambiguity of the measures,
we showed the behaviour of the measures through numerical experiments (Section 6.4).
In this section, we discuss the effect of the IoU thresholding on the estimation stability.

We here show the case of $K=1$ for a simple example.
Let $r$ be the IoU value of the top-1 moment for the true unobservable ground truth and
$\hat{r}$ be that for the noisy ground truth.
Under a Gaussian noise model $\hat{r}=r + \epsilon$, where $\epsilon \sim N(0, \gamma^2)$,
noisy IoU $\hat{r}$ also obeys a normal distribution $\hat{r} \sim N(r, \gamma^2)$.
The expected difference between the true and observed AxIoU@1 (\ie bias) is obtained as $\mathbb{E}_{\hat{r}}[r - \hat{r}] = r - \mathbb{E}_{\hat{r}}[\hat{r}]$.
Hence, AxIoU@1 is unbiased (\ie $\mathbb{E}_{\hat{r}}[r - \hat{r}]=0$) because $\hat{r}$ is an unbiased estimator of $r$ (\ie $\mathbb{E}_{\hat{r}}[\hat{r}]=r$).
The variance of AxIoU@1 is exactly that of $\hat{r}$ (\ie $\mathbb{V}[\hat{r}]=\gamma^2$).

On the other hand, the bias of R@$1,\theta$ can be obtained as
\begin{align*}
  \mathbb{E}_{\hat{r}}[\mathds{1}\{\hat{r} \geq \theta\}-\mathds{1}\{r \geq \theta\}] &= \mathbb{E}_{\hat{r}}[\mathds{1}\{\hat{r} \geq \theta\}] - \mathds{1}\{r \geq \theta\} \\
  &= P(\hat{r} \geq \theta) - \mathds{1}\{r \geq \theta \}.
\end{align*}
If $\theta \leq r$ holds, because the true R@$1,\theta$ is one,
the bias is then $P(\hat{r} \geq \theta) - 1 = -P(\hat{r}<\theta)$.
If $\theta > r$ holds, the bias is $P(\hat{r} \geq \theta)$.
Therefore, R@$1,\theta$ is statistically biased; that is, it has the error even in the expectation.
Because $\mathds{1}\{\hat{r}\geq \theta\}$ obeys the Bernoulli distribution $\mathrm{Bern}(P(\hat{r} \geq \theta))$,
The variance of R@$1,\theta$ is $P(\hat{r} \geq \theta)P(\hat{r} < \theta)$, which depends on $\theta$ and $\gamma$.
Figure~\ref{fig:variance-example} shows the theoretical (squared) bias, variance and mean squared error (MSE) of AxIoU@1 and R@$1,\theta$ for different $\theta$ and $\gamma$ under $r=0.5$.
We can observe that both AxIoU@1 and R@$1,\theta$ have large estimation errors (\ie MSE) when noise level $\gamma$ is large.
In addition to this, R@$1,\theta$ suffers from a severe error even with small $\gamma$, particularly when $\theta$ is close to $r=0.5$.
This is an undesirable property because we often need to discriminate competitive VMR methods and thus to use $\theta$ around the boundary, which leads to estimation errors under label noise.

\clearpage
{\small
\bibliographystyle{ieee_fullname}
\bibliography{egbib}
}

\end{document}